\documentclass[letterpaper, 10 pt, conference]{ieeeconf}  

\usepackage{xcolor}
\usepackage{amsmath,amsfonts}
\usepackage{nccmath} 
\usepackage{amssymb}
\usepackage{algorithm}
\usepackage{algorithmic}
\usepackage{graphicx}
\usepackage{url}
\usepackage{multirow}
\usepackage{balance}
\usepackage{tabularx}

\IEEEoverridecommandlockouts                              

\title{\LARGE \bf
A Real-Sim-Real (RSR) Loop
Framework for Generalizable \\Robotic Policy Transfer
}

\author{Yuxuan Xu$^{2+}$, Shiyu Wang$^{3+}$, Jinhao Huang$^{1}$, Wenhao Zhao$^{4}$, Yufei Jia$^{1}$ Zike Yan$^{1}$, Weibin Gu$^{1}$, \\
Lu Shi$^{1*}$, and Guyue Zhou$^{1*}$
\thanks{The authors are with $^{1}$ Tsinghua University, $^2$ Zhejiang University, $^3$ Beijing Institute of Technology, and $^4$The Hong Kong University of Science and Technology (GZ), $^{+}$ for co-first and $^{*}$ for corresponding authors.}%
\thanks{We gratefully acknowledge the support of Wuxi Research Institute of Applied Technologies, Tsinghua University under Grant 20242001120 and the Shuimu Scholarship of Tsinghua University. Any opinions, findings and conclusions or recommendations expressed in this material are those of the authors and do not necessarily reflect the views of the funding agencies.
}
}

\begin{document}

\maketitle
\thispagestyle{empty}
\pagestyle{empty}

\begin{abstract}
The sim-to-real gap remains a critical challenge in robotics, hindering the deployment of algorithms trained in simulation to real-world systems. We propose a flexible Real–Sim–Real (RSR) framework whose central contribution is an information-theoretic cost function that explicitly accounts for sim-to-real discrepancies. This cost balances two objectives: completing the task and steering the policy to collect real-world samples that are maximally informative for improving transfer. It can be integrated seamlessly into existing reinforcement learning algorithms (e.g., PPO, SAC) and ensures a balanced exploration of critical regions in the real domain. The framework treats differentiable simulation as optional: when a differentiable simulator is available, the collected informative data can also be used to tune simulator parameters.
We implement the framework with the MuJoCo MJX platform and demonstrate its generality by evaluating on both manipulation tasks with a 6-DOF robotic arm and locomotion tasks on a legged robot. Empirical results show that our RSR loop yields more efficient data acquisition and substantially improves task performance in real-world that achieves a smoother sim-to-real transfer. The code and Appendix content is in the anonymized GitHub repo at~\url{https://anonymous.4open.science/r/RSR-C917}.   
\end{abstract}

\section{Introduction}
Simulation has become a cornerstone of robot learning, enabling safe, cost-effective, and rapid development of control policies without risking physical hardware~\cite{control2023}. Despite these advantages, policies trained in simulation often fail to transfer reliably to real robots due to the sim-to-real (sim2real) gap—discrepancies between simulated and real-world conditions that arise from imperfect modeling of physics, sensor noise, and environmental uncertainty~\cite{hofer2021sim2real,ge2024bridging}. Consequently, simulation-trained policies may exhibit degraded performance or unsafe behaviors when deployed in real environments~\cite{salvato2021s2r}.

Various approaches have been proposed to address these challenges~\cite{zhao2020s2r}. One of the most widely used techniques is Domain Randomization (DR)~\cite{tobin2017DR}. DR enhances generalization by increasing the diversity of the simulated environment, training algorithms across a wide range of parameter variations. Although effective, DR often requires manual selection of parameters to randomize, lacks scalability as the complexity of the problem grows, and operates as an open-loop approach that does not incorporate feedback from real-world data. Another promising avenue lies in domain adaptation~\cite{bousmalis2018DA}, where the sim2real problem is framed as a transfer learning task. In this paradigm, the simulation domain (source domain) and the real world domain (target domain) are bridged through feature alignment techniques, such as adversarial methods~\cite{ganin2016adversial} or reconstruction-based approaches~\cite{bousmalis2016domain}. Despite their success in computer vision, adapting these methods to robotics poses unique challenges, particularly in extracting and leveraging invariant features for domain matching and transfer. Additionally, neural-augmented simulation approaches~\cite{golemo2018LSTM} employ recurrent models such as LSTMs to predict and compensate for discrepancies between simulated and real-world dynamics, while recent imitation-based pipelines, such as VideoMimic~\cite{allshire2025videomimic}, reconstruct environments and humanoid motions from videos and leverage DAgger-style~\cite{ross2011dagger} aggregation to improve sim-to-real transfer across both simulated and real humanoid platforms. However, these approaches often require large amounts of real-world data and are sensitive to noise, limiting their robustness.

Recent work has demonstrated the effectiveness of leveraging real-world data to reduce the sim-to-real gap~\cite{maddukuri2025simandreal,heiden2021neuralsim}. For instance, Liu et al.~\cite{liu2020rsr} propose a Real-Sim-Real (RSR) transfer method in which a task-relevant simulator is constructed using semantic information during a real-to-sim phase, and the resulting policy is then deployed directly on the real robot without requiring additional real-world data. BayesSim~\cite{ramos2019bayessim} further enhances domain randomization by incorporating real-world observations to infer a posterior distribution over simulation parameters. In a related direction, Liu et al.~\cite{liu2024diffrender} integrate vision-based foundation models with robotic tasks, enabling robot pose reconstruction from video via forward kinematics and gradient-based optimization. To better organize the related work, we categorize the existing methods that leverage real-world data for sim-to-real transfer into three groups: \textbf{(1)} a growing body of work focuses on \textbf{explicit simulator or world-model optimization} using real data~\cite{torne2024reconciling}. Differentiable simulators~\cite{newbury2024review} enable simulation parameters to be fine-tuned through gradient-based optimization to better match real-world dynamics~\cite{qiao2021articulated}. For example, GradSim~\cite{murthy2020gradsim} defines key physical parameters such as mass and control forces, compares simulated states with real observations (e.g., videos), and updates simulator parameters via backpropagation. Heeg et al.~\cite{heeg2024quadvisual} further establish a fully differentiable pipeline that links pixel observations, state estimation, and reinforcement learning, significantly improving policy performance. More recently, EmbodieDreamer~\cite{wang2025embodiedreamer} introduces neural modules such as PhysAligner and VisAligner to optimize physical parameters and visual appearance from real observations, while PIN-WM~\cite{li2025pin-wm} proposes a physics-informed world model for efficient end-to-end identification of 3D interaction dynamics in non-prehensile manipulation. These methods primarily emphasize improving simulator fidelity or learning neural approximations of dynamics and observation mappings. \textbf{(2)} Another line of research addresses sim-to-real discrepancies by \textbf{calibrating actuator dynamics or compensating execution errors} using real-robot data. Bridging the Sim-to-Real Gap for Athletic Loco-Manipulation~\cite{feybridging} trains an Unsupervised Actuator Network (UAN) that maps histories of joint position and velocity errors to corrective torques, enabling the calibration of complex actuator transmission mechanisms. Similarly, ASAP~\cite{he2025asap} introduces a residual action policy trained in simulation, whose outputs are added to nominal actions to compensate for simulator imperfections during real-world execution. \textbf{(3)} In parallel, \textbf{iterative system identification} approaches seek to adapt simulation parameters online. Dynamics as Prompts~\cite{zhang2025context} formulates system identification as an in-context learning problem, using a causal transformer that takes interaction histories as prompts and predicts environment parameters via next-token prediction, enabling on-the-fly simulator adaptation through alternating rollouts in simulation and the real world. 

However, these approaches face several challenges. First, many existing methods overlook potential biases in real-world data collection, which can lead to suboptimal tuning results. The assumption that real-world data accurately represents the robot’s domain may not always hold, especially if the data is collected in a non-representative manner or fails to cover the regions of interest for the policy. Second, some methods rely solely on visual appearance or video data for tuning. While this approach can provide useful information, it is highly sensitive to factors such as lighting conditions, camera angles, and sensor calibration, which require rigorous experimental setups. Additionally, visual data primarily captures information about the pose of the robot, but it often fails to provide insights into critical dynamic variables like velocity, acceleration, or thrust, which are essential for accurate simulation and control. Finally, many of these approaches are limited to specific types of robots or environments, reducing their generalizability.

To address these limitations, we introduce a flexible Real–Sim–Real (RSR) framework driven by an information-theoretic cost function. Rather than random data collection, our framework uses this cost to explicitly encode sim-to-real gap considerations, guiding the policy to collect real-world data that is maximally informative for improving transfer while still fulfilling the task objective. Related ideas have been explored in MetaMVUC~\cite{gilles2024metamvuc}, while only focused on visual uncertainty. The cost integrates naturally with standard reinforcement learning algorithms, such as PPO~\cite{PPO2023} and SAC~\cite{SAC2024}, and can incorporate different types of modalities to shape exploration toward sim-to-real–critical regions. Although a differentiable simulator may be used when available, it is not required, making the framework compatible with a wide range of simulation pipelines. We implement our approach on the MuJoCo MJX engine~\cite{MuJoCo2023} and demonstrate its effectiveness through experiments on both robotic manipulation with a 6-DOF arm and locomotion tasks on a legged robot. Finally, we discuss the impact of including visual terms in the cost, current limitations, and future research directions.

\begin{figure*}[ht!]
    \centering
    \includegraphics[trim = {3cm 5.5cm 3cm 5cm},clip, width=0.85\textwidth]{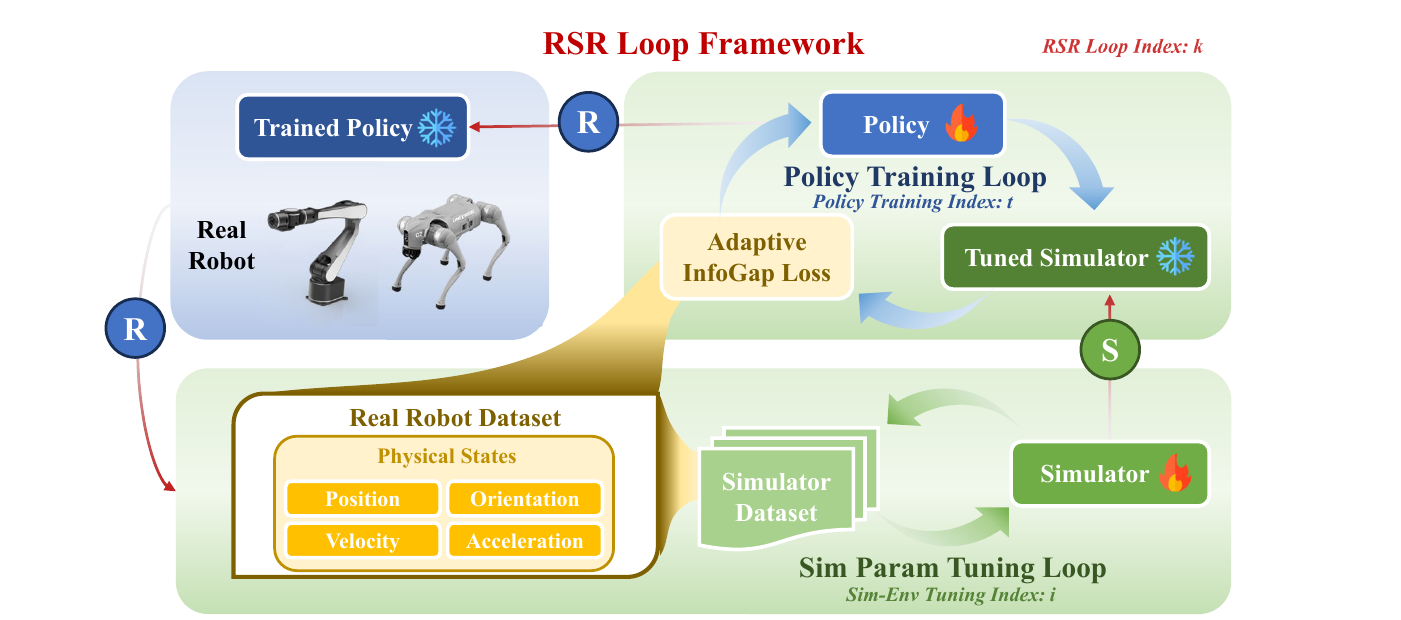}
    {\caption{Overview of the proposed RSR (Real–Sim–Real) loop framework. The red outer loop (arrows labeled “R”) is the main Real–Sim–Real cycle: a simulation-trained policy is deployed on the real robot, the collected data are used to update the InfoGap cost (and, optionally, the simulator), and a new policy is then trained in simulation. The blue inner loop is the policy-training loop, which uses the current simulator and the adaptive InfoGap loss. The green inner loop is the optional simulator-parameter-tuning loop (indexed by i). Together, these loops iteratively improve both the policy and, when applicable, the simulator.}}
    \label{fig:Overview}
\vspace{-10pt}
\end{figure*}
\section{Preliminary}
\subsection{Information-Theoretic Preliminaries}
\label{sec:appendix-B}

We briefly summarize the information-theoretic tools used in this work, focusing on density estimation and distributional divergence measures.

\subsubsection{Kernel Density Estimation (KDE)}
Kernel Density Estimation (KDE)~\cite{chen2017KDE} is a non-parametric method for estimating a probability density function from a finite set of samples $\{x_i\}_{i=1}^n$. Given a kernel function $K(\cdot)$ and bandwidth $h$, the estimated density is
\begin{equation}
\hat{p}(x) = \frac{1}{n h} \sum_{i=1}^{n} K\!\left( \frac{x - x_i}{h} \right).
\end{equation}
The bandwidth $h$ controls the smoothness of the estimated distribution.

\subsubsection{Distribution Divergence Measures}
The Kullback--Leibler (KL) divergence quantifies the discrepancy between two probability distributions $p$ and $q$:
\begin{equation}
\mathrm{KL}(p \,\|\, q) = \sum_{x} p(x)\, \log \frac{p(x)}{q(x)} .
\end{equation}

The Wasserstein distance (Earth Mover's Distance) measures the minimal transport cost required to transform one distribution into another. The $\beta$-order Wasserstein distance between $p$ and $q$ is defined as
\begin{equation}
W_{\beta}(p, q) =
\left(
\inf_{\gamma \in \Gamma(p,q)}
\int_{\mathcal{X} \times \mathcal{X}} d(x,y)^{\beta} \, d\gamma(x,y)
\right)^{\!\!1/\beta},
\end{equation}
where $\Gamma(p,q)$ denotes the set of joint distributions with marginals $p$ and $q$, and $d(\cdot,\cdot)$ is the ground metric.

\section{Method}
\subsection{System Overview}
Figure~\ref{fig:Overview} illustrates the proposed Real–Sim–Real (RSR) loop framework. 
At its core, the framework is driven by an information-theoretic cost that explicitly encodes sim-to-real gap considerations and guides the policy to collect real-world data that is maximally informative for improving transfer, while still achieving the task objective. Rather than relying on random data collection or simulator-specific assumptions, this cost directly shapes policy behavior and is compatible with general simulation environments.

The RSR loop proceeds iteratively as illustrated in the Algorithm~\ref{alg:main}: a policy trained in simulation is deployed on the real robot to collect new interaction data, the data is then used to construct the InfoGap cost that guides the next round of policy optimization in simulation.
When a differentiable simulator is available, real-world data can additionally be used to tune environment parameters via gradient-based optimization; however, such tuning is optional and not required for the framework to function.
Even without simulator parameter updates, the information-driven cost alone promotes purposeful exploration and improves sim-to-real robustness across iterations (which is validated in the experiments).
In the following, we will outline the details of each component of the proposed structure. 

%
\begin{algorithm}
\caption{The RSR Loop Framework}
\begin{algorithmic}[1]\label{alg:main}
\STATE \textbf{Initialize:} Initial policy $\pi_0$, environment parameters $\theta_0$, and initial real dataset $\mathcal{D}_{real}^0$ collected from the deployment of $\pi_0$ on the real robot.
\WHILE{not converged, at iteration $k$}
    \STATE \textbf{(Optional) Step 1:} \textbf{Environment Parameter Tuning}
    \STATE \quad Initialize $\theta_{0,k-1} = \theta_{k-1}$, compute the corresponding loss function $\mathcal{L}(\theta_{i,k-1})$ 
    with the real dataset $\mathcal{D}_{real}^k$and update the differentiable simulation's parameters as in~\eqref{eq:parameter} $\theta_{i,k-1}$ using gradient descent 
    $$
    \theta_{i,k-1} \gets \theta_{i-1,k-1} - \alpha \nabla_\theta \mathcal{L}(\theta_{i-1,k-1})
    $$
    until they converge and assign $\theta_k = \theta_{i^{\star},k-1}$.
    \STATE \textbf{Step 2:} \textbf{Policy Training with the InfoGap Cost}
    \STATE \quad Using the updated environment $ \theta_k $, perform reinforcement learning training for the policy $\pi_k$ with the adaptive InfoGap cost~\eqref{eq:AdaptGap} included:
    $$
    \mathcal{L}(\pi_k^t) = \mathbb{E}_{s_t, a_t \sim \pi_k} \left[ r_t + \gamma \mathbb{V}_{\pi_k}(s_{t+1}) - \mathbb{V}_{\pi_k}(s_t) \right]+\mathcal{L}_{sr}^k
    $$
    \STATE \textbf{Step 3:} Deploy the updated policy $\pi_k$ on the real robot and collect new real-world data $\mathcal{D}_{real}^{k+1}$. Repeat the process for subsequent iterations.
\ENDWHILE
\end{algorithmic}
\end{algorithm}

\subsection{Simulation-Env Parameter Tuning Process}
When a differentiable simulator is available, real-world data can be used to align the simulation environment with real-world dynamics by tuning environment parameters $\theta$

Concretely, at iteration $k$, we deploy the policy $\pi^{(k-1)}$ trained in simulation onto the real robot and collect a trajectory of state--action--next-state transitions,
\begin{equation*}
\mathcal{D}_{\mathrm{real}} =
\left\{(s_t, a_t, s_{t+1})\right\}_{t=1}^{T},
\end{equation*}
which captures the real-world system dynamics induced by the current policy. Using the same state--action pairs $\{(s_t, a_t)\}_{t=1}^{T}$, we roll out the simulator parameterized by $\theta$ to obtain the corresponding simulated transitions,
\begin{equation*}
\mathcal{D}_{\mathrm{sim}}(\theta) =
\left\{(s_t, a_t, \hat{s}_{t+1}(\theta))\right\}_{t=1}^{T}.
\end{equation*}

We then define a physical discrepancy loss that measures the mismatch between real and simulated state transitions. When the simulator is differentiable, the environment parameters can be optimized using gradient-based updates as:
\begin{equation}\label{eq:parameter}
\theta \leftarrow \theta - \alpha \nabla_{\theta}\bigg(
\frac{1}{T}
\sum_{t=1}^{T}
\ell\big(s_{t+1}, \hat{s}_{t+1}(\theta)\big)\bigg).
\end{equation}


\subsection{Adaptive InfoGap Loss construction}
As discussed earlier, leveraging real-world data has become an increasingly common and effective paradigm for sim-to-real transfer. However, how such data are collected strongly affects their effectiveness. Common data collection strategies~\cite{collection2023} include random sampling, grid-based sampling, and trajectory-based sampling. Random sampling~\cite{ransampling2024} selects states independently, often leading to uneven coverage and missing task-relevant regions. Grid-based sampling~\cite{grid2022} provides systematic coverage but becomes computationally prohibitive in high-dimensional spaces. While trajectory-based sampling—commonly used in robotic learning—collects data along policy rollouts and is naturally aligned with task execution. However, trajectory-based data collection is inherently biased toward regions explored by the current policy, limiting exposure to unmodeled or underrepresented dynamics. To mitigate this, we balance the estimation of the real domain with task completion throughout the iterations, ensuring that the policy optimizes for task performance while exploring underrepresented regions of the real world.

\begin{figure}[h]
\centering\includegraphics[width=0.4\textwidth]{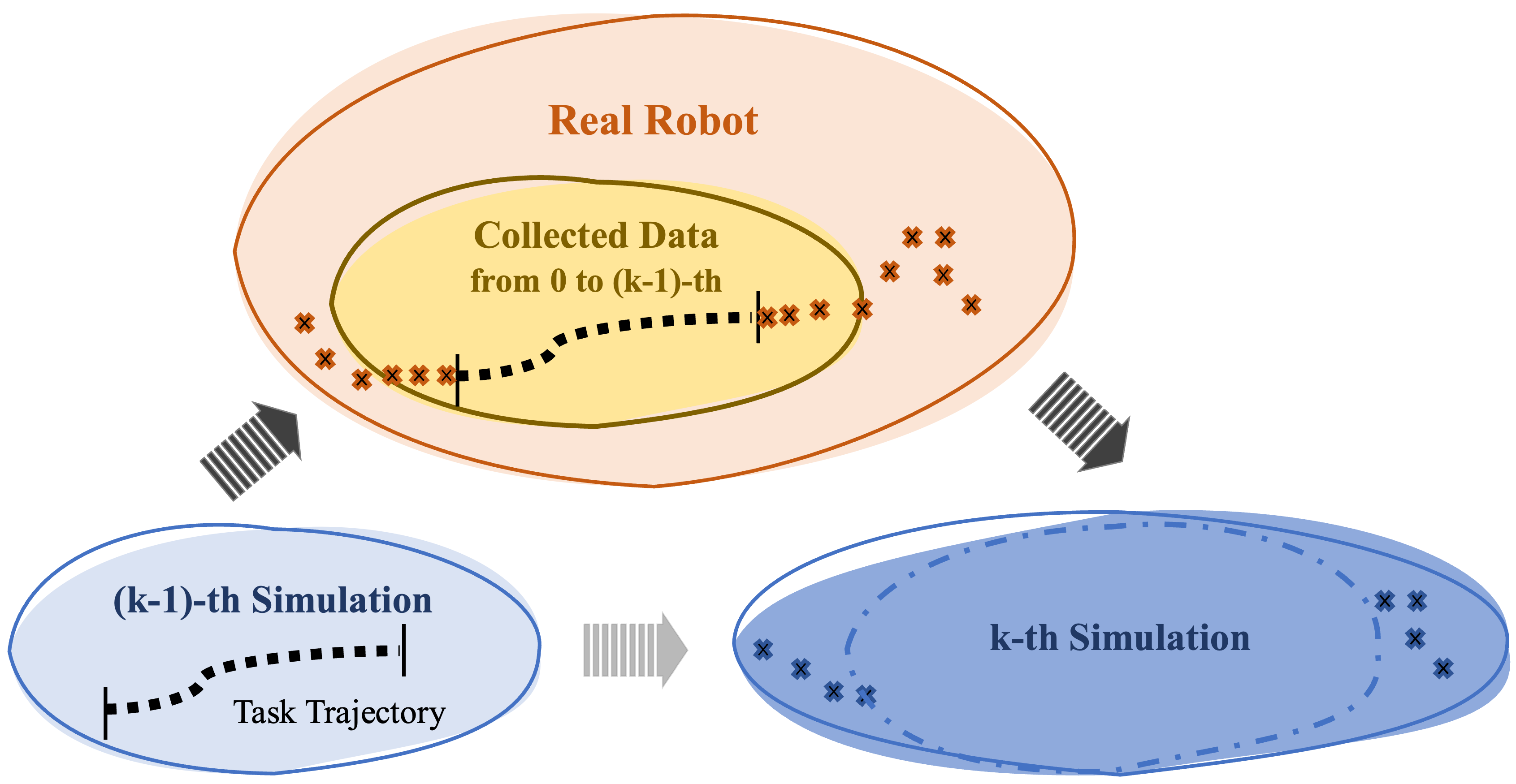
}
\vspace{-5pt}
    \caption{The process of bridging the sim-to-real gap in robot training. When the discrepancy between the simulation (blue domain) and real robot (orange domain) is large, the policy prioritizes collecting informative data (marked as crosses) from the real domain to better characterize its properties other than the task trajectory (black dashed line). 
    }
    \label{fig:loss}
    \vspace{-10pt}
\end{figure}
The core idea as shown in Fig.~\ref{fig:loss} is that when the discrepancy between the simulation environment and the real robot is substantial, the policy prioritizes collecting more informative data that better characterizes the real domain properties rather than focusing solely on task completion. This is because a policy trained under such a significant sim-to-real gap is unlikely to perform well on the real robot. As the gap narrows and the simulator parameters converge towards the real domain, this term in the loss function diminishes, allowing the policy to shift its focus to task completion. 
Specifically, this adaptive InfoGap cost and the corresponding minimization problem to determine the action at timestep $t$ during the 
$k$-th sim-to-real tuning iteration is constructed as
$$a_{k,t} = \text{argmin} \enspace \mathcal{L}(a_{k,t}) = \mathcal{L}_{task}(a_t)+\mathcal{L}_{sr}(a_{k,t})\enspace ,$$ where $\mathcal{L}_{task}$ is the nominal cost $\mathcal{L}_{task} = \mathbb{E}[(r_t + \gamma V(s_{t+1}) - V(s_t)) \nabla_{\pi} \log \pi(a_t | s_t)]$ for task completion and
\begin{equation}\label{eq:AdaptGap} 
\begin{medsize}
    \mathcal{L}_{sr} = -\text{KL}\left(\hat{p}(\mathcal{D}_{real}^k)\parallel \hat{p}(\mathcal{D}^{k-1}_{sim})\right) \cdot W_{\beta}\left(\hat{p}(\mathcal{D}_{sim}^{k}+\mathcal{D}_t),\hat{p}(\mathcal{D}_{sim}^{k})\right) \enspace.
\end{medsize}
\end{equation}
%
The meanings of each notation are listed below and more details can be found at the Appendix in the supplemental material:
\begin{itemize}
    \item $\mathcal{D}_{real}^k$: the data set of length $M$ $\{s_m,a_m=\pi_{k-1}(s_m),s_{m+1}^{real}\}_0^M$ collected from the actual robot with the policy $\pi_{k-1}$ trained from the last iteration $k-1$. 
    
    \item $\mathcal{D}_{sim}^{k-1}$: the data set of the same input but propagated in the untuned simulator $\{s_m,a_m,s_{m+1}^{sim}\}_0^M$.

    \item $\mathcal{D}_{sim}^k$: the dataset propagated in this tuned new simulator $\{s_m,a_m,\hat{s}_{m+1}^{sim}\}_0^M$.


   \item $\mathcal{D}_t$: the data pair generated by the current input $a_{k, t}$.

    \item $\hat{p}(\{\mathcal{D}\})$ represents the distribution estimation function from data samples set $\mathcal{D}$.
    
    \item $\text{KL}(p,q)$ and $W_{\beta}$ represents the KL divergence and Wasserstein distance, seperately.
\end{itemize}
%

At the beginning of each training iteration \( k \) in the simulation, the first term, \( \text{KL}\left(\hat{p}(\mathcal{D}_{real}^k) \parallel \hat{p}(\mathcal{D}^{k-1}_{sim})\right) \), remains constant throughout the training process and represents the sim-to-real gap at the current iteration. Meanwhile, a specific data pair \( \mathcal{D}_t \), influenced by the policy's action \( a_t \) during training, affects the second term, \( W_{\beta}\left(\hat{p}(\mathcal{D}_{sim}^{k,t}+\mathcal{D}_t), \hat{p}(\mathcal{D}_{sim}^{k,t})\right) \). This term encourages the policy to generate actions that produce more informative data—data that exhibit larger discrepancies from the simulated distribution while being closer to the real-world distribution.
Therefore, the cost function is designed to encourage actions that are more informative and exploratory, enabling the collection of data that fully captures the distribution or characteristics of the real domain, thereby mitigating data bias when addressing the sim-to-real gap. When the sim-to-real gap becomes negligible (i.e., $\text{KL}\left(\hat{p}(\mathcal{D}_{real}^k)\parallel \hat{p}(\mathcal{D}^{k-1}_{sim})\right)$ approaches zero), this term $\mathcal{L}_{sr}$ in the cost function diminishes, leaving only the task completion cost $\mathcal{L}_{task}$ to guide the actions. Ultimately, this approach yields a policy capable of completing the task effectively while minimizing the sim-to-real gap.

{Note that unlike offline random or grid-based sampling, our setting requires data to be collected through physically executable robot rollouts. Independent random-state queries are generally unavailable on real hardware because the robot cannot be reset arbitrarily to every state, while grid-based coverage scales exponentially with the state-action dimension and is therefore impractical for contact-rich systems. In addition, our method does not fine-tune the policy on real-world data: all policy optimization is performed in simulation. Real-world trajectories are used only to estimate the sim-to-real discrepancy, construct the InfoGap term, and optionally update simulator parameters. This design preserves the safety and sample-efficiency advantages of simulation-based RL while using limited real interaction to guide informative exploration.}
\section{Experiments}
In this section, we present a set of experiments to evaluate the effectiveness and generality of the proposed RSR loop framework across different robotic platforms and task settings.
Specifically, we consider (i) block-pushing tasks with a 6-DOF robotic arm, (ii) T-shaped object manipulation, and (iii) velocity tracking with a quadrupedal robot, {and (iv) payload-perturbed velocity tracking with the quadrupedal robot using SAC.}
{For all manipulation tasks (i and ii), the framework incorporates simulation parameter adaptation enabled by a differentiable simulator, allowing physical parameters to be refined using real-world data.
In contrast, the quadrupedal experiments (iii and iv) employ only the proposed information-gap cost without explicit simulator parameter tuning, demonstrating that our framework remains effective with general, non-differentiable simulators.}
{Together, these experiments highlight the versatility of the RSR loop across heterogeneous robotic systems and RL algorithms.}
Finally, we discuss the potential of extending the framework by incorporating a visual loss component into the sim-to-real adaptation process. 

{Simulations are conducted in DISCOVERSE~\cite{discoverse2024}, using PPO and SAC with domain randomization.
The resulting PPO+DR and SAC+DR policies, evaluated before any RSR update, serve as the domain-randomization baselines.
Real-world evaluations are conducted on a 6-DOF AIRBOT Play robotic arm and a Unitree Go2 quadrupedal robot, with the task-specific action, observation, reward, and domain-randomization settings detailed in Appendix}


\subsection{Robotic Arm Block-Pushing Task}



In this experiment, the robot is required to push the block to random target points within the workspace. 
{We first train a PPO policy with domain randomization in simulation. Its rollout in simulation is marked as ``simulated trajectory'', while its direct deployment on the real robot is marked as ``PPO+DR''. We then apply successive RSR iterations, with the resulting real-world trajectories denoted as $n$-th RSR Trajectory.}
{As shown in Table~\ref{tab:kl_divergence_block}, the PPO+DR baseline exhibits large KL divergences in the block pose between simulated and real-world trajectories, indicating that domain randomization alone does not fully eliminate the discrepancy between simulation and real-world execution.}
This indicates a significant discrepancy between the simulation-trained policy and the real-world environment.
{After successive RSR iterations, the KL divergences are substantially reduced overall, showing that the updated simulator and policy become better aligned with real-world behavior.}
This demonstrates the effectiveness of the designed InfoGap loss. 
\hspace{-20pt}
\begin{table}[htbp]
\centering
\caption{Distribution Deviation Results in the Block-Pushing Task}
\label{tab:kl_divergence_block}
\begin{tabular}{|c|c|c|}
\hline
\textbf{Stage} & \textbf{KL Divergence ($x$)} & \textbf{KL Divergence ($y$)} \\ \hline
PPO+DR       & 16.3509                           & 36.3168                           \\ \hline
1st\_RSR       & 1.6903                            & 3.4206                            \\ \hline
2nd\_RSR       & 5.0462                            & 2.3946                            \\ \hline
3rd\_RSR       & 0.9739                            & 0.8384                            \\ \hline
\end{tabular}
\vspace{-12pt}
\end{table}

\begin{figure}[h]
    \centering
    \includegraphics[width=0.4\textwidth]{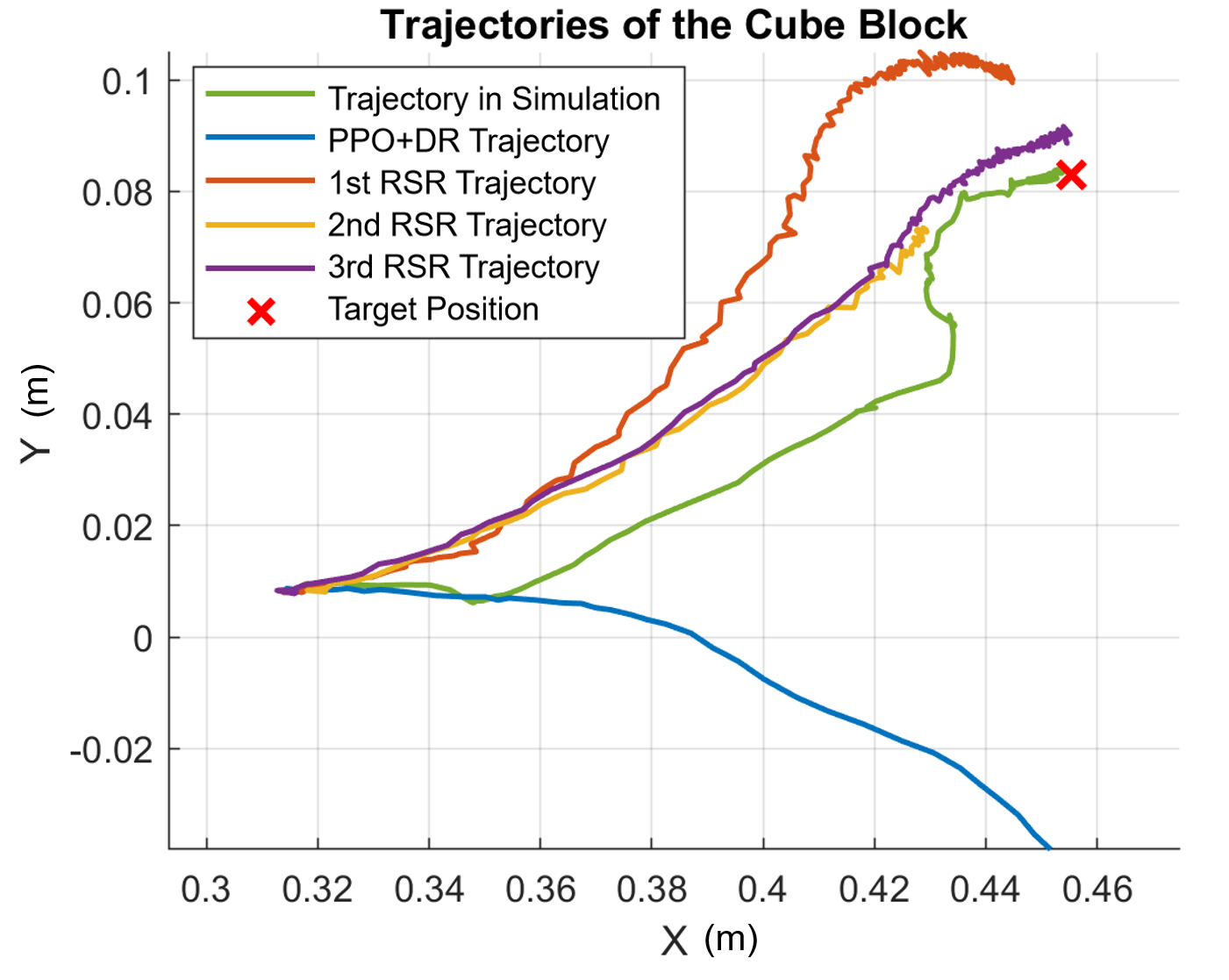}
    \caption{Real-world block-pushing trajectories across different iterations.}
    \label{fig:single_result}
\end{figure}
The improvement in trajectory alignment is further visualized in Fig.~\ref{fig:single_result}, which shows the block's trajectories at different training stages averaged over three independent runs.
{The PPO+DR baseline trajectory (blue) deviates substantially from the simulated trajectory and fails to reach the target.}
The low friction between the cube and the desk as well as the misalignment between the gripper and the cube causes the gripper to slip along the cube's surface instead of effectively pushing it, resulting in the cube moving in an unintended direction.
{In contrast, the trajectories obtained after successive RSR iterations (orange, yellow, and purple) show an overall improvement in accuracy.
By 3rd RSR (purple), the real-world trajectory closely matches the simulated trajectory, and the block successfully reaches the target position (red cross ``$\times$'').}
These results highlight the effectiveness of the RSR approach in closing the sim-to-real gap. By iteratively refining the simulation model and retraining the policy, we achieve a substantial reduction in trajectory divergence and significantly improve the real-world task performance. 

\subsection{Robotic Arm T-Shaped Object Pushing Task}
\begin{figure}[hb]
\vspace{-5pt}
    \centering
    \includegraphics[width=0.45\textwidth]{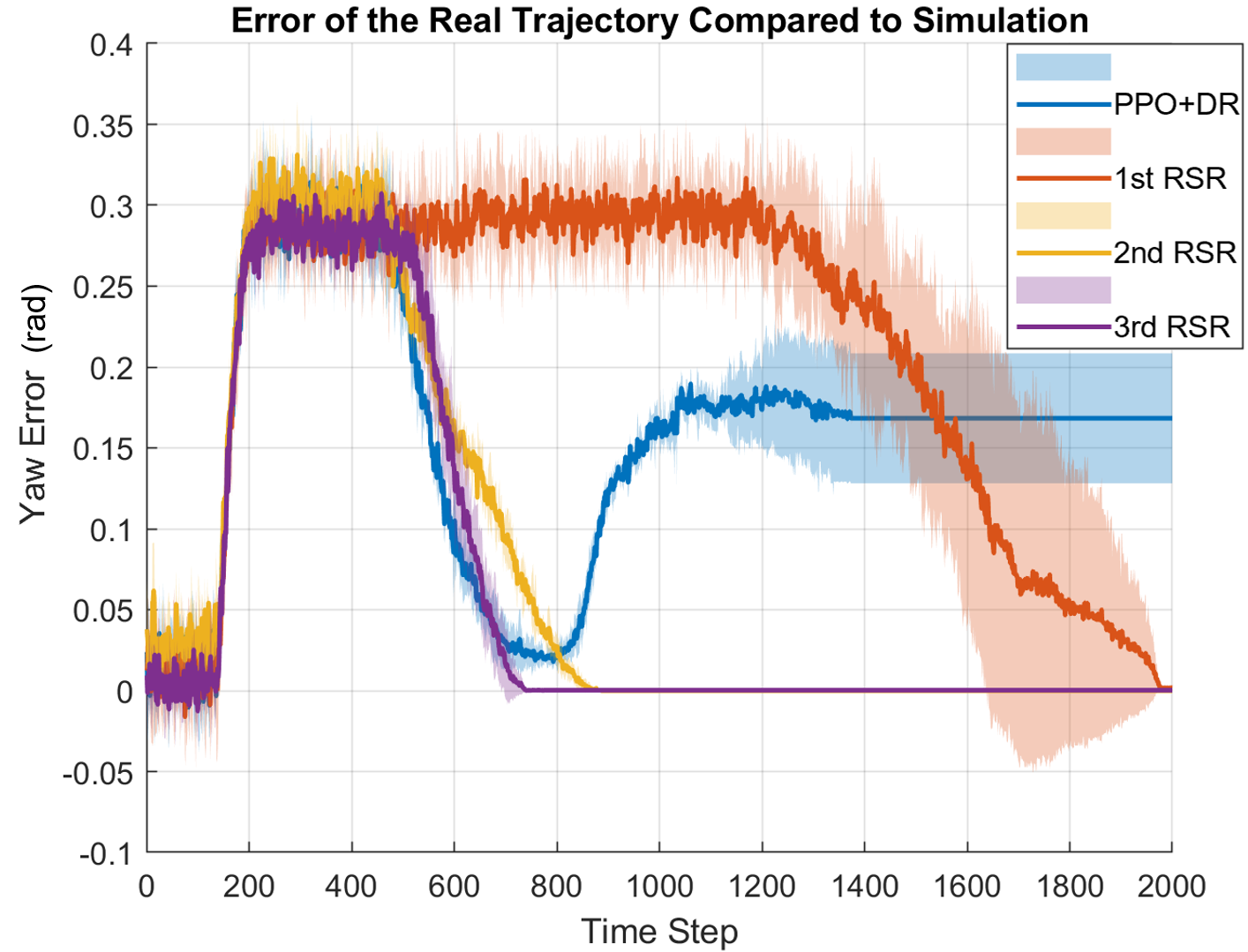}
    \caption{1-$\sigma$ bounds of real trajectories for the yaw angle in the T-shaped block pushing trials for different iterations, where the shaded area represents the bounds.}
    \label{fig:t_result}
\end{figure}

In this variant of the block-pushing task, the robot is required to push a T-shaped object to a specified target position. This experiment provides a more challenging scenario due to the object’s geometry and the increased difficulty of manipulating irregularly shaped objects. We did the similar experiments as we evaluated in the cube block. Three trials were conducted and the results are shown in Fig.~\ref{fig:t_result}.
{The figure shows that the yaw error is reduced across the RSR iterations. The PPO+DR baseline (blue) exhibits a large yaw error, while 1st RSR (red) reduces the error but remains unstable. 2nd RSR (yellow) and 3rd RSR (purple) achieve substantially lower errors, showing that the RSR updates further improve sim-to-real alignment beyond the domain-randomized PPO baseline.}

\subsection{Quadrupedal Robot Circular Velocity-Tracking Task}
In this experiment, the RSR framework is evaluated on a Unitree Go2 quadrupedal robot performing velocity tracking tasks. The robot is commanded to execute a constant velocity profile $[v_x, v_y, \omega_z] = [0.7, 0.0, -0.5]$ (m/s, m/s, rad/s), which induces forward motion with a steady right turn, resulting in a circular trajectory. The same command is executed in simulation and on the real robot across multiple RSR iterations.

Table~\ref{tab:kl_divergence} reports the KL divergences of each dimension computed at each iteration, which serve as inputs to the information-gap cost for the next RSR cycle. Starting from the initial deployment, large distributional discrepancies are observed, indicating substantial mismatch between simulated and real-world dynamics. Similar to the manipulation tasks, as the RSR loop progresses, the KL terms used to construct the information-gap cost decrease steadily, reflecting that the policy is being guided toward regions of the state–action space where the simulation and real-world behaviors become increasingly consistent.

\begin{table}[h!]
\centering
\caption{Distribution Deviation of the Circular-Tracking Task}
\label{tab:kl_divergence}
    \begin{tabular}{|c|c|c|c|}
    \hline
    \multirow{2}{*}{\textbf{Stage}} & \multicolumn{3}{c|}{\textbf{KL Divergence}}\\
    \cline{2-4}
     & $v_x$ (m/s) & $v_y$ (m/s) & $\omega_z$ (rad/s)  \\ \hline
    PPO+DR       & 48.2941                           & 5.1624               & 56.7724                      \\ \hline
    1st\_RSR       & 38.3659                           & 3.5171               & 47.5421                      \\ \hline
    2nd\_RSR       & 9.2554                            & 2.6966               & 19.5236                      \\ \hline
    3rd\_RSR       & 4.3988                            & 1.9778               & 3.9715                      \\ \hline
    \end{tabular}
    \vspace{-15pt}
\end{table}

\begin{figure}[h]
\hspace{-10pt}
    \centering
    \includegraphics[width=0.35\textwidth]{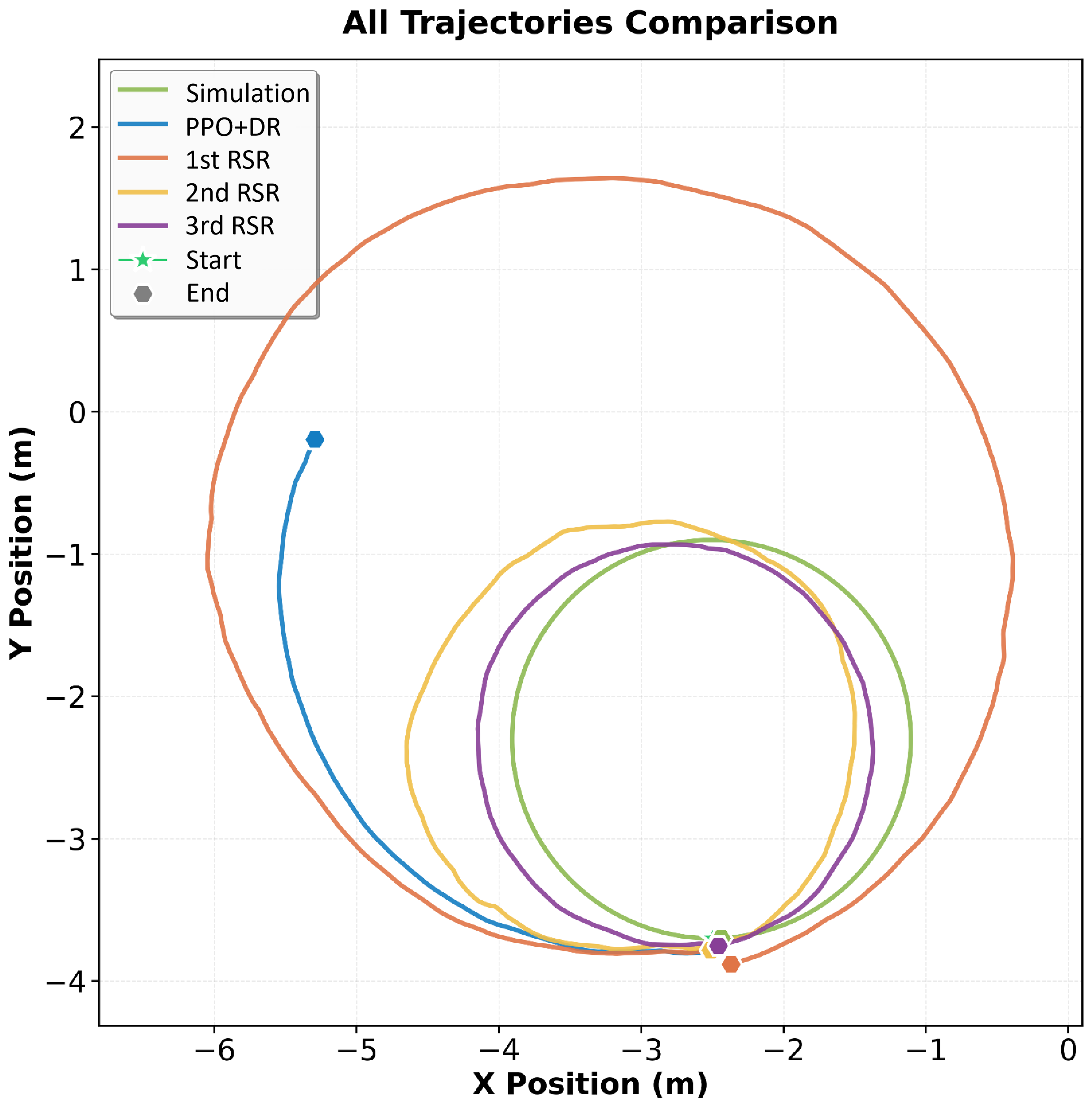}
    \caption{Real-world legged trajectories across different iterations.}
    \label{fig:legged}
\end{figure}

{The simulated trajectory serves as the reference circular path. As shown in Fig.~\ref{fig:legged}, the PPO+DR baseline, despite being trained with domain randomization in simulation, exhibits substantial deviation from the reference when directly deployed on the real robot. This remaining discrepancy reflects sim-to-real mismatch caused by unmodeled contact dynamics, actuator delays, and ground interaction effects.}
As RSR iterations proceed, the real-world trajectories progressively converge toward the simulation reference, with later iterations closely reproducing the intended circular motion. We evaluate tracking accuracy using the root mean square error (RMSE) between the commanded and measured body-frame velocities.
{Table~\ref{tab:velocity_error} shows an overall reduction in velocity-tracking errors across RSR iterations.}
Notably, the yaw-rate tracking error is reduced by more than a factor of five, which plays a critical role in stabilizing the robot’s turning behavior and directly contributes to the improved circular trajectory. Similar, though less pronounced, improvements are observed for the forward and lateral velocity components.

{Overall, compared with the PPO+DR baseline, successive RSR iterations substantially improve tracking performance in legged locomotion.}
{Importantly, this experiment introduces only the proposed information-gap cost beyond the PPO+DR baseline, without explicit simulator parameter tuning, thereby isolating its contribution and demonstrating its applicability to contact-rich robotic systems beyond manipulation.}

\begin{table}[h!]
\centering
\caption{Velocity Tracking RMSE of the Quadrupedal Robot}
\label{tab:velocity_error}
\begin{tabular}{|c|c|c|c|}
\hline
\textbf{Stage} & \textbf{$v_x$ (m/s)} & \textbf{$v_y$ (m/s)} & \textbf{$\omega_z$ (rad/s)} \\ \hline
PPO+DR       & 0.194                & 0.077                & 0.334                       \\ \hline
1st\_RSR       & 0.147                &  0.050                & 0.295                       \\ \hline
2nd\_RSR       & 0.067                & 0.068                & 0.086                       \\ \hline
3rd\_RSR       & 0.051                & 0.045                & 0.052                       \\ \hline
\end{tabular}
\vspace{-10pt}
\end{table}

\subsection{Quadrupedal Robot Payload-Tracking Task}

{To further evaluate the generality of RSR across RL algorithms and dynamics variations, we integrate RSR with SAC and evaluate the Unitree Go2 under three payload conditions: no payload, a centered 2.5-kg payload, and the same payload placed 0.15~m forward of the torso center. All conditions execute the same 20-s velocity-tracking sequence, consisting of $[v_x,v_y,\omega_z]=[0.3,0,0]$, $[0.7,0,0]$, $[0.5,0,0.8]$, and $[0.5,0,-0.8]$, and start independently from the same SAC+DR policy.
{
RSR is agnostic to the underlying RL algorithm. For SAC, we use the standard actor objective as the task objective:
}
\begin{equation}
\mathcal{L}^{\mathrm{SAC}}_{\mathrm{task}}
\!=\!
\mathbb{E}_{s_t\sim\mathcal{B},\,a_t\sim\pi}
\!\left[
\alpha \log \pi(a_t|s_t)
\!-\!
\min_{j=1,2} Q_j(s_t,a_t)
\!\right]\!.
\end{equation}
{
The overall policy objective remains
$\mathcal{L}^{\mathrm{SAC}}_{\mathrm{task}}+\mathcal{L}_{\mathrm{sr}}$,
where $\mathcal{L}_{\mathrm{sr}}$ is the InfoGap cost defined in Eq.~(2).
The critic and entropy-temperature updates follow standard SAC.
}
For a controlled comparison, we additionally evaluate direct real-world fine-tuning by inserting the same real-world transitions into the SAC replay buffer.
}

\begin{table}[h]
\centering
\caption{Distribution Deviation across RSR Iterations under Different Payload Conditions}
\label{tab:sac_payload_kl}
\begin{tabular}{|c|c|c|c|c|}
\hline
\textbf{Payload} & \textbf{Stage} & $v_x$ (m/s) & $v_y$ (m/s) & $\omega_z$ (rad/s) \\ \hline
\multirow{4}{*}{None}
& SAC+DR  & 44.8173 & 4.9362 & 51.2847 \\ \cline{2-5}
& 1st RSR & 30.9648 & 4.0837 & 35.6724 \\ \cline{2-5}
& 2nd RSR & 17.3826 & 2.9145 & 19.4861 \\ \cline{2-5}
& 3rd RSR & 8.7519  & 1.7286 & 8.9375  \\ \hline

\multirow{4}{*}{Centered}
& SAC+DR  & 52.4618 & 6.2743 & 63.9052 \\ \cline{2-5}
& 1st RSR & 37.5831 & 4.8619 & 44.7168 \\ \cline{2-5}
& 2nd RSR & 21.0467 & 3.4296 & 24.5732 \\ \cline{2-5}
& 3rd RSR & 10.6843 & 2.1938 & 11.7629 \\ \hline

\multirow{4}{*}{Front-biased}
& SAC+DR  & 61.7384 & 7.4196 & 74.5261 \\ \cline{2-5}
& 1st RSR & 45.2617 & 5.7463 & 53.1849 \\ \cline{2-5}
& 2nd RSR & 26.7385 & 4.1378 & 29.8614 \\ \cline{2-5}
& 3rd RSR & 13.4926 & 2.7641 & 14.6857 \\ \hline
\end{tabular}
\end{table}

{
The SAC+DR policy already exhibits robustness to the payload variations, while larger residual discrepancies remain under the front-biased payload. Across all three conditions, successive RSR iterations substantially reduce the KL divergence between simulated and real velocity distributions.
}

{ We further compare RSR with direct real-world fine-tuning using the same real-world data. Specifically, at each adaptation round $k$, we reuse the real-world transitions $\mathcal{D}_{\mathrm{real}}^k$ collected for RSR, assign rewards using the same task reward function, and insert them directly into the SAC replay buffer for continued standard SAC training, without the InfoGap cost. We denote this baseline as SAC+Real Replay. Thus, both methods start from the same SAC+DR policy and use exactly the same real-world transition sets at each round, differing only in how these limited real-world data are utilized. }

\begin{figure}[h!]
    \centering
    \includegraphics[width=0.5\textwidth]{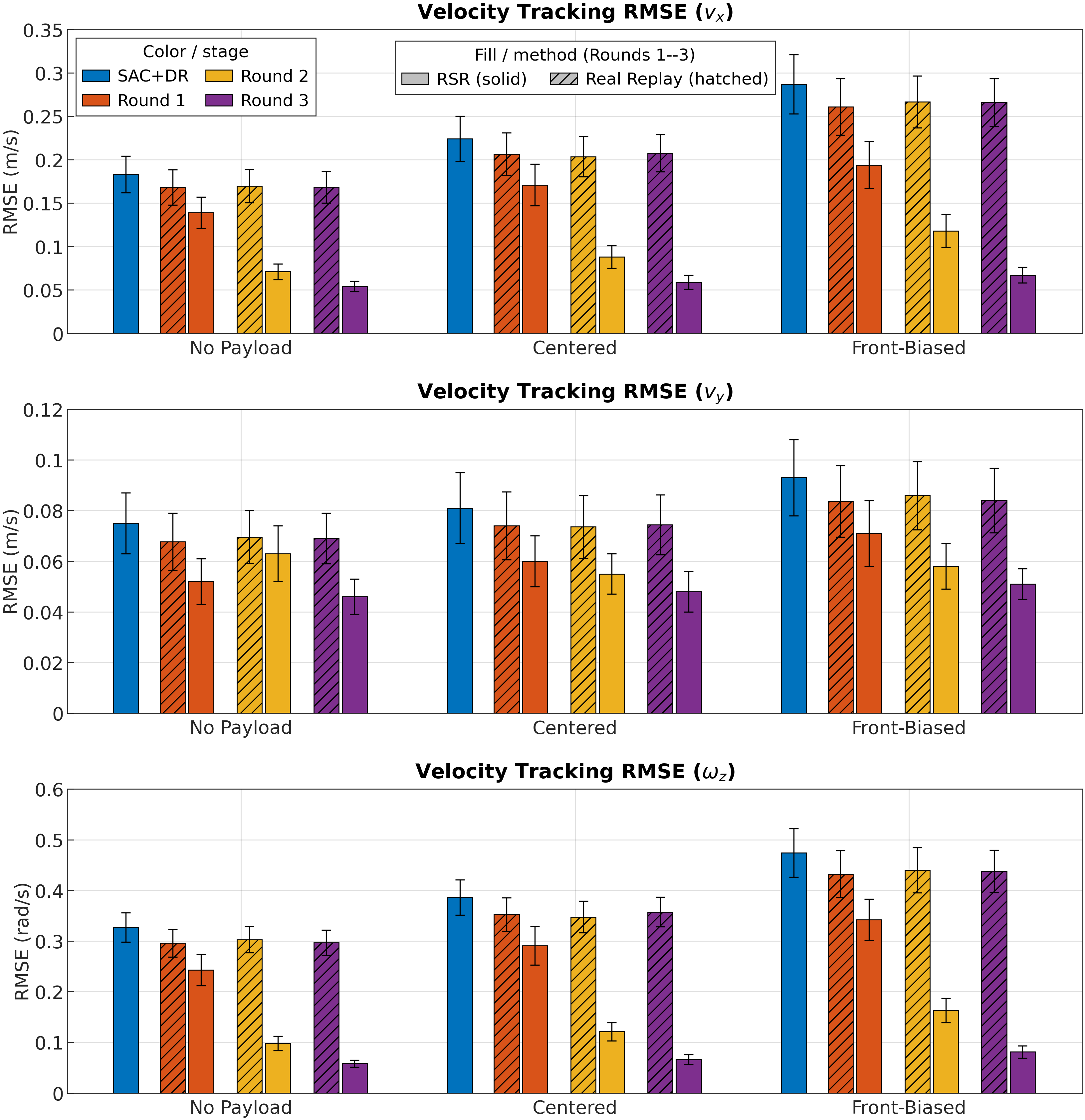}
    \caption{Velocity-tracking RMSE under different payload conditions across successive RSR and SAC+Real Replay iterations. Both methods use the same real-world transition sets at each adaptation round.}
    \label{fig:payload_rmse}
\end{figure}

{
As shown in Fig.~\ref{fig:payload_rmse}, directly incorporating the same real-world transitions into the SAC replay buffer results in performance nearly identical to SAC+DR, whereas successive RSR iterations consistently reduce the velocity-tracking error, with particularly clear improvements under the front-biased payload. Notably, each adaptation round uses only approximately one minute of real-world interaction, which is negligible compared with the amount of simulated experience used for SAC training. Consequently, directly replaying these sparse real-world transitions has little influence on standard SAC optimization. In contrast, RSR converts the same limited real-world data into the InfoGap cost to guide subsequent simulation-based training, leading to substantial and consistent improvements. These results demonstrate that RSR can effectively exploit very limited real-world data without requiring direct real-world policy fine-tuning.}

\subsection{Discussion on Incorporating Visual Loss}
Other than the main evaluation, we explored the possibility of incorporating a visual loss component into the {policy training process.} Specifically, we computed the visual loss using the Structural Similarity Index (SSIM~\cite{sara2019SSIM}) between Gaussian-rendered simulation images and real-world images.
{We selected SSIM as an initial lightweight visual metric because its local structural comparison is sensitive to object boundaries and contour misalignment, which are particularly relevant in our rendered manipulation scenes.}
{We use it as a lightweight visual-consistency baseline to examine whether improving image-level structural alignment can benefit simulator adaptation.}

\vspace{-13pt}
\begin{figure}[h]
    \centering
    \includegraphics[trim = {0cm 0.2cm 0cm 0cm}, clip, width=0.5\textwidth]{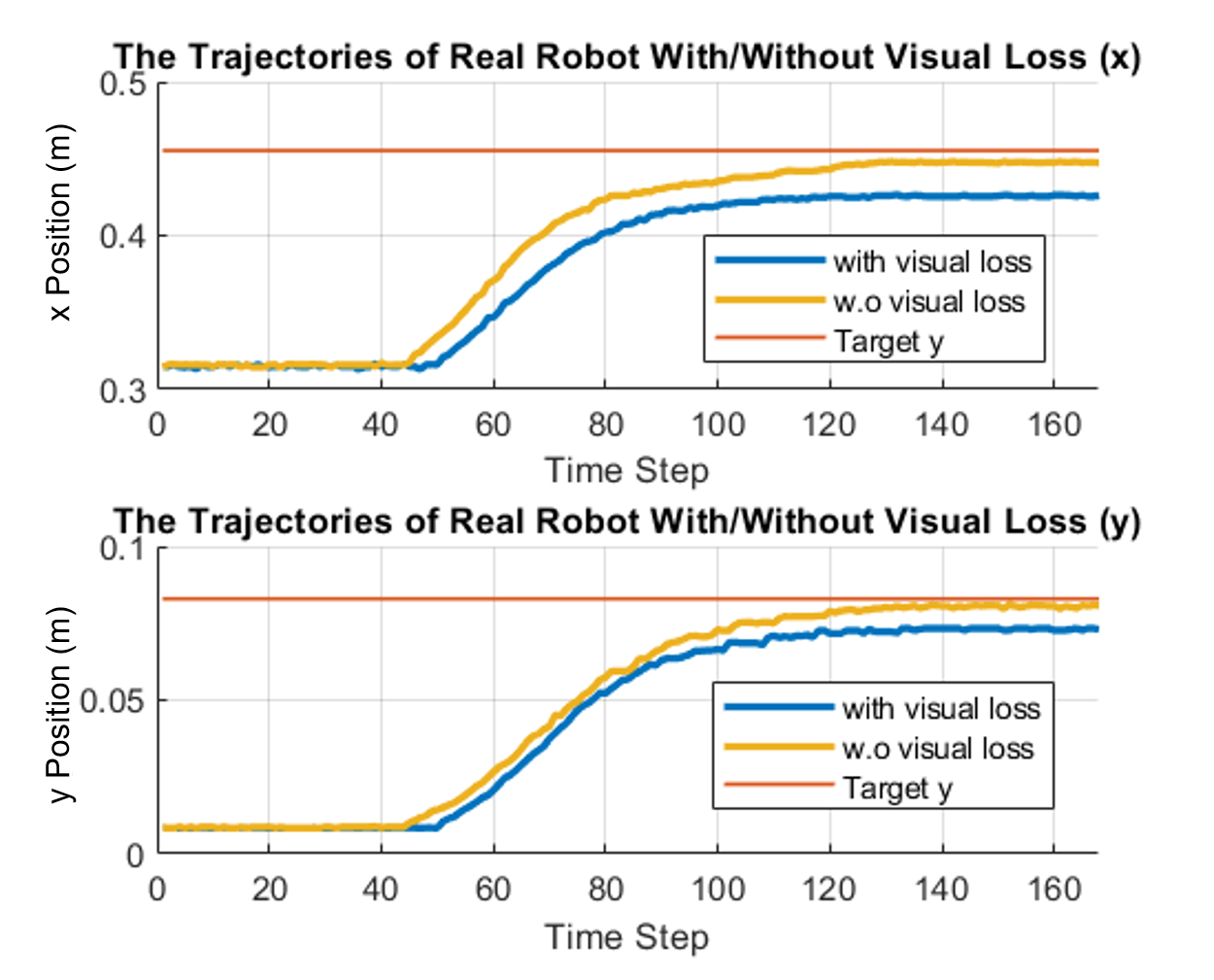}
    \caption{Trajectories of real-robot deployment with and without visual loss.}
    \label{fig:visual}
    \vspace{-10pt}
\end{figure}
However, as shown in Fig.~\ref{fig:visual}, the results indicate that incorporating the visual loss in this manner does not lead to improved performance. In fact, it appears to introduce instability in the adaptation process. One possible reason for this is that while SSIM effectively captures high-level structural similarity, it does not directly correlate with the physical parameters that affect policy learning. As a result, using SSIM-based visual loss as a weight may amplify discrepancies in ways that are not meaningful for improving real-world task performance. Additionally, since the visual differences can be caused by lighting conditions, reflections, and camera noise—factors that do not directly impact the underlying dynamics—forcing the simulator to match real images too closely might introduce misleading optimization directions. Another critical limitation we observed is the computational cost associated with rendering high-fidelity images for environment parameter tuning. Real-time rendering with physically realistic lighting and textures requires significant GPU resources, making it impractical for iterative training. Given these challenges, we have chosen not to include the visual loss in our current pipeline. 



\section{Discussion and Conclusion}
In this work, we propose an RSR loop framework for sim-to-real transfer that leverages real-world data through an information-gap cost, with optional simulation parameter adaptation when a differentiable simulator is available. Through experiments on robotic manipulation and quadrupedal locomotion, we validate that the proposed cost enables smoother sim-to-real transfer across iterations and improves real-world performance. We further examine the use of a visual loss for sim-to-real adaptation. The observed gains are modest under the current experimental conditions, likely due to computational constraints and the sensitivity of visual alignment to domain-specific factors, which motivates further investigation in future work. These observations, 
point to clear directions for future work, including more efficient visual representations, improved scalability, and extension to platforms such as aerial robots operating in highly dynamic environments.

\bibliographystyle{IEEEtran}
\bibliography{IEEEabrv,IEEEexample}

\newpage

\balance
\end{document}